%% file: neurips_2026.tex
\documentclass{article}

\PassOptionsToPackage{numbers, compress}{natbib}
\usepackage[preprint]{neurips_2026}
\usepackage{amsmath}
\usepackage{xspace}

\usepackage[utf8]{inputenc} % allow utf-8 input
\usepackage[T1]{fontenc}    % use 8-bit T1 fonts
\usepackage{hyperref}       % hyperlinks
\usepackage{url}            % simple URL typesetting
\usepackage{booktabs}       % professional-quality tables
\usepackage{amsfonts}       % blackboard math symbols
\usepackage{nicefrac}       % compact symbols for 1/2, etc.
\usepackage{microtype}      % microtypography
\usepackage{xcolor}         % colors
\usepackage{cleveref}
\usepackage{graphicx} 
\usepackage{multirow}
\usepackage{wrapfig}
\usepackage{enumitem}

\newcommand{\methodname}{FLAT\xspace}

\title{\methodname: Feedforward Latent Triangle Splatting for Geometrically Accurate Scene Generation}

\author{%
  \textbf{Orest Kupyn}\textsuperscript{1,2}\thanks{Work done during an internship at Google.} \quad
  \textbf{Goutam Bhat}\textsuperscript{1} \quad
  \textbf{Philipp Henzler}\textsuperscript{1} \\
  \textbf{Fabian Manhardt}\textsuperscript{1} \quad
  \textbf{Christian Rupprecht}\textsuperscript{1,2} \quad
  \textbf{Federico Tombari}\textsuperscript{1,3} \\[2mm]
  \normalfont
  \textsuperscript{1}Google Research \\
  \textsuperscript{2}University of Oxford, Visual Geometry Group \\
  \textsuperscript{3}TU Munich
}

\begin{document}

\maketitle

\input{sections/0-abstract}
\input{sections/1-introduction}
\input{sections/2-related_work}
\input{sections/3-method}
\input{sections/4-evaluation}
\input{sections/5-conclusion}

\section*{Acknowledgments}
C.R. is funded by the European Union (ERC, Volute, 101222037). Views and opinions expressed are however those of the author(s) only and do not necessarily reflect those of the European Union or the European Research Council. Neither the European Union nor the granting authority can be held responsible for them. We thank Oleg Gordiichuk for the help with the mobile rendering demo.

\bibliographystyle{plainnat}
\bibliography{main}

\clearpage
\appendix
\input{sections/7-appendix}

% \newpage
% \input{checklist.tex}

\end{document}

%% file: sections/0-abstract.tex
\begin{abstract}

Generating explorable 3D scenes from a single image requires strong generative priors and accurate geometric representations suitable for downstream use. Current video diffusion models offer high-quality generation and implicitly encode multi-view geometric structure in latent space. However, existing feedforward latent scene decoders typically output volumetric 3D Gaussians that lack a well-defined surface, limiting their use in simulation or standard graphics pipelines. This motivates decoding \emph{surface-aligned} primitives that are not only renderable but also closer to explicit geometric assets. We ask whether compressed video diffusion latents can be mapped directly to explicit surface primitives in a single pass. To this end, we introduce \methodname and, for the first time, show that triangle splats can be decoded directly from video diffusion latents. Compared with decoding 3D Gaussians, predicting flat primitives is notoriously more challenging due to high sensitivity to primitive orientations, oftentimes leading to poor gradient flow. \methodname solves with two key ingredients: a ray-centered rotation parameterization for triangle regression and a novel \textit{product window function} that improves gradient flow during differentiable triangle rendering. On standard benchmarks, \methodname achieves significantly better geometric accuracy while maintaining competitive visual quality compared to state-of-the-art feedforward baselines. We further show that a lightweight test-time refinement step converts the predicted triangle soup into a fully opaque, game-engine-ready representation that supports real-time rendering. By evaluating 3DGS, 2DGS, and triangle splatting variants under an identical training setup, we provide the first systematic analysis of representation tradeoffs in feedforward scene generation. The project page is available at \url{https://flat-splat.github.io}.
\end{abstract}

%% file: sections/1-introduction.tex
\section{Introduction}

Creating explorable 3D environments is a challenging problem, with applications in mixed reality \cite{yuan2026immersegen}, robotics simulation \cite{gao2026dreamdojo, kim2023neuralfield, yang2024holodeck}, game asset creation \cite{xu2024sketch2scene}, and autonomous driving \cite{xscene, mao2025dreamdrive}. These applications require not only visually plausible content but also geometrically accurate, physically grounded scene representations that capture 3D layout, surface structure, and scale to support novel view synthesis, physical simulation, and interaction. The challenge is compounded when only a single image or text caption is available as input. The scene is under-determined: depth is ambiguous, and occluded surfaces are uncovered as the camera moves. Generating a complete explorable 3D scene from a single image, therefore, requires strong geometric and generative priors~\cite{wonderland,lyra}. 

Recent advances in video diffusion models  \cite{wan, Sora, veo3_reasoning, hunyuanvideo, moviegen} offer a viable path towards this goal. These models learn rich priors and implicit 3D world understanding from internet-scale data. Nevertheless, video diffusion models alone cannot support interactive scene exploration due to high render time. Furthermore, they cannot ensure multi-view consistency.
% Thus, explicit 3D representations remain important to overcome these limitations.
A number of approaches thus follow a generate-then-optimize paradigm~\cite{viewcrafter,worldstereo,cat3d} wherein a 3D Gaussian Splatting \cite{3dgs} or NeRF \cite{nerf} representation is optimized to fit frames generated by the video model. This enables real-time rendering, but introduces substantial computational overhead due to the per-scene optimization.

%Yet, generating pixels is not equivalent to 3D scene: no explicit geometry or guarantee of multi-view consistency. Furthermore, running inference of a video diffusion model at render time is expensive. Explicit 3D representations remain necessary.
% The challenge is compounded when only a single image or text caption is available as input. The scene is under-determined: occluded regions are invisible, disoccluded surfaces appear as the camera moves, and depth is ambiguous. Generating a complete, explorable 3D scene from a single image requires a strong generative prior, motivating the use of video generation models.

% A natural way to combine video diffusion models with explicit geometry is to follow a generate-then-optimize paradigm. Camera-controlled video diffusion models such as ViewCrafter \cite{viewcrafter}, WorldStereo \cite{worldstereo}, or multi-view generation models \cite{cat3d} synthesize multiple views of a scene, then fit a 3D Gaussian Splatting \cite{3dgs} or NeRF \cite{nerf} representation to these frames. This produces navigable scenes of reasonable quality, but at a very high cost. In particular, running a large diffusion model followed by  per-scene optimization is computationally very expensive. As the video latent already encodes geometric structure during the denoising process, decoding latent into pixels only to re-estimate geometry from scratch discards all this information, thus wasting much compute. This motivates a more direct approach to decode 3D scene parameters directly from the video latents.

\begin{figure*}[t]
    \centering
    \includegraphics[width=0.95\textwidth]{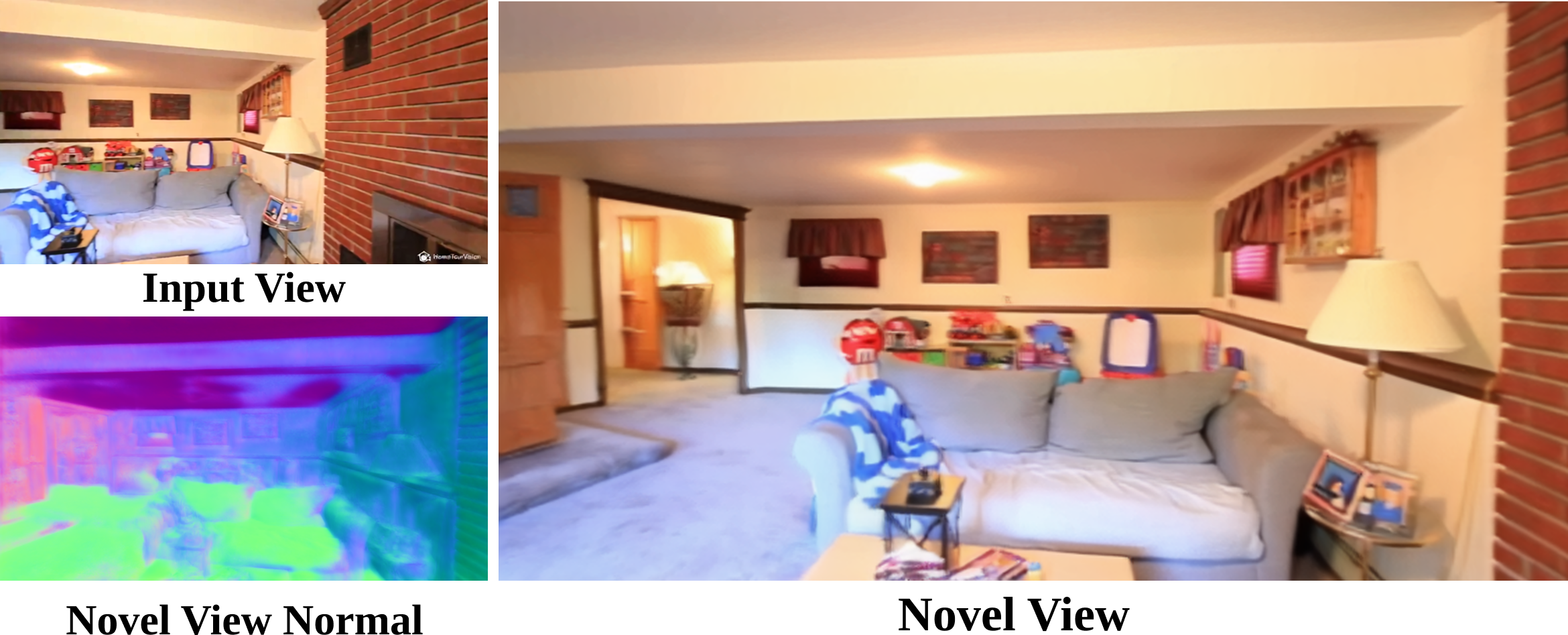}
    \caption{\methodname regress soft triangles directly from video diffusion latent, enabling geometrically accurate and high fidelity scene generation.}
    \vspace{-0.45cm}
    \label{fig:main_page}
\end{figure*}

Wonderland \cite{wonderland}, Lyra \cite{lyra}, and Generative Gaussian Splatting \cite{ggs} demonstrate that scene parameters can be regressed directly from video latents using lightweight decoders on top of frozen video diffusion models. By minimizing photometric error between original and rendered images, these approaches can generate high-quality, explorable 3D scenes. However, all these methods are restricted to generating only 3D Gaussians as output; these are volumetric, semi-transparent blobs that are well-suited to training scene decoders via differentiable rendering. Yet these same properties make them unsuitable for most graphics engines, which rely on opaque surface representations such as triangles or meshes~\cite{held2025meshsplatting}. While there are approaches to extract meshes from a Gaussian representation~\cite{wolf2024gs2mesh, guedon2025milo}, these often require complex post-processing and cannot produce satisfactory results. 

Directly optimizing opaque triangle or mesh-based representations for the scene, however, is difficult due to the non-differentiability of the rendering process \cite{liu2019soft, chen2019learning}. Soft triangle representations~\cite{triangle_splatting, held2025meshsplatting} can overcome this issue to enable per-scene optimization. Unfortunately, training feedforward triangle splatting decoder presents itself with further challenges. Exemplary, directly regressing vertices can easily result in degenerate solutions. Unlike volumetric Gaussians, incorrectly oriented flat triangles contribute negligibly to rendered images, yielding poor gradient supervision, especially early in training. Together, these issues make stable feedforward prediction of non-volumetric primitives an open problem, requiring careful choices in both parameterization and differentiable rendering.

We introduce \methodname, a feedforward model that directly predicts semi-opaque triangle-splatting primitives \cite{triangle_splatting, held2025meshsplatting} from the latent space of a frozen video diffusion model in a single forward pass. Given an input image, \methodname produces a geometrically accurate, physically grounded scene representation supervised by depth and normals. We enable efficient feedforward triangle decoding with two technical ingredients. First, we formulate a stable parameterization for flat primitives: each decoder token predicts a ray-centered triangle defined by a constrained Cholesky-style shape transform and residual rotations around a ray-aligned frame, avoiding degenerate triangles and unstable world-space orientations. Second, we introduce a modified window function that improves gradient flow across primitive boundaries during differentiable triangle rendering. Feedforward triangle model produces significantly more accurate geometry and on-par visual quality with volumetric variants \Cref{fig:main_page}. For compatibility with standard graphics pipelines and game engines, we also provide an optional lightweight refinement step that converts the semi-transparent feedforward prediction into fully opaque triangles. We also train 3DGS \cite{3dgs} and 2DGS variants \cite{2dgs} under identical conditions, enabling direct comparison of the representations. Our contributions are as follows:
\begin{itemize}[left=0pt]
    \item We show for the first time that explicit, \textit{non-volumetric} surface primitives can be decoded directly from compressed video diffusion latents in a single forward pass, and formulate the previously underexplored problem of how to parameterize and train feedforward flat-primitive decoding.
    \item We introduce the key ingredients that make this practical: a ray-centered local triangle parameterization with constrained Cholesky-style shape, residual orientation prediction around a ray-aligned frame, and a novel product window function that improves gradient flow and stabilizes training.
    \item We introduce \methodname, a feedforward pipeline from a single image to a game-engine-compatible format, and provide the first systematic comparison of 3DGS, 2DGS, and triangle splatting under identical latent decoding conditions, characterizing tradeoffs among rendering quality, geometric accuracy, and downstream mesh compatibility.
\end{itemize}

%% file: sections/2-related_work.tex
\section{Related Work}

\paragraph{Novel View Synthesis and Scene Generation}

3D scene generation methods can be grouped into several categories. Early works train multi-view generation models \cite{liu2023zero, shi2023mvdream, wang2023imagedream, cat3d} to expand the viewset and then reconstruct an explicit 3D representation \cite{nerf, 3dgs}. Recently, ViewCrafter \cite{viewcrafter} has extended the generate-then-optimize paradigm to video diffusion, generating large, dense views from a point-cloud-conditioned video model. WorldStereo \cite{worldstereo} adds explicit memory and 3D consistency optimization to video diffusion, enabling large-scale viewpoint generation. Yet all these methods require a complex two-stage pipeline with an expensive scene-optimization step, which limits scalability and computational efficiency. Recent feedforward novel view synthesis models \cite{charatan2024pixelsplat, chen2024mvsplat, szymanowicz2024splatter, ye2025yonosplat, xu2025depthsplat, da3, zhang2025flare, gslrm, longlrm} predict explicit 3D scene parameters, typically 3DGS, from RGB images. This streamlines the 3D scene generation pipeline by replacing complex scene optimization with a lightweight feedforward model. Yet such a pipeline discards all the intermediate features generated by a multi-billion-parameter video model only to re-estimate scene geometry from pixels.

Alternatively, geometry-free novel view synthesis methods completely omit the explicit prediction of 3D scene parameters. LVSM \cite{jin2024lvsm} directly maps input images to novel-view outputs, completely eliminating intermediate scene representations. LagerNVS incorporates features from geometry foundation models \cite{wang2025vggt}, showing the effectiveness of 3D-aware latent features for geometry-free generation. Recently, Genie \cite{bruce2024genie} released a model that generates novel views in near-real time with high 3D consistency. Yet it requires the entire model to run for every new view rendered, demanding substantial computational resources. Thus, for many tasks, explicit 3D representations remain crucial.

Wonderland \cite{wonderland} extends the video diffusion model to 3D by training a 3DGS decoder directly from the latent space, enabling efficient single-stage 3D scene generation. Yet the task is highly complex: the decoder must infer scene geometry, appearance, and depth solely from rendering losses in a frozen, compressed latent space, while being guided by often imperfect cameras. Generative Gaussian Splatting \cite{ggs} demonstrates that the highest quality is achieved with a second-stage scene optimization that starts from feedforward latent model predictions, which, yet again, increases the pipeline complexity. Lyra \cite{lyra} improves the quality of the feedforward latent 3DGS model by incorporating multi-view video latents generated from a small set of preset trajectories. This allows bypassing supervision for noisy and complex trajectories but limits the scale and diversity of the generated scenes. All of the latent feedforward scene generation models are based on 3DGS. In contrast, \methodname explores a non-volumetric triangle representation for accurate scene geometry and direct compatibility with rendering engines. We also enable generation of diverse camera trajectories without post-scene optimization by improving the decoder architecture and camera pose guidance.

\paragraph{3D Scene Representations}

NeRF-style \cite{nerf} volumetric representations established a powerful framework for view synthesis, but their rendering cost motivated a shift towards more efficient explicit scene parameterizations. 3D Gaussian Splatting \cite{3dgs} showed that collections of anisotropic 3D Gaussians enable high-quality real-time rendering. However, while Gaussian-based representations are flexible and efficient, they do not always provide the surface regularity, geometric precision, or structural control that some downstream tasks require. Thus, various modifications and alternatives of 3D Gaussians have been explored \cite{hamdi2024ges, taktasheva20253d, huang2025deformable, chen2024beyond}, including 2D Gaussian splatting \cite{2dgs} for improved geometric accuracy. Recently, completely different representations, such as smooth 3D convexes \cite{held20253d} or radiance foams \cite{govindarajan2025radiant}, have been proposed. Triangle Splatting \cite{triangle_splatting} introduces differentiable rendering of soft triangles -- the most classical primitive in computer graphics. MeshSplatting \cite{held2025meshsplatting} further extends this line of work by enabling connectivity, allowing for differentiable mesh optimization. However, extending feedforward scene generation beyond Gaussians is challenging: non-volumetric representations have compact gradients and require precise orientation. \methodname addresses these issues, showing that triangles can be predicted directly from video latent, which in turn enables efficient and flexible scene generation with strong geometric accuracy and direct compatibility with modern rendering engines.

%% file: sections/3-method.tex
\section{Method}

\begin{figure*}[t]
    \centering
    \includegraphics[width=\textwidth]{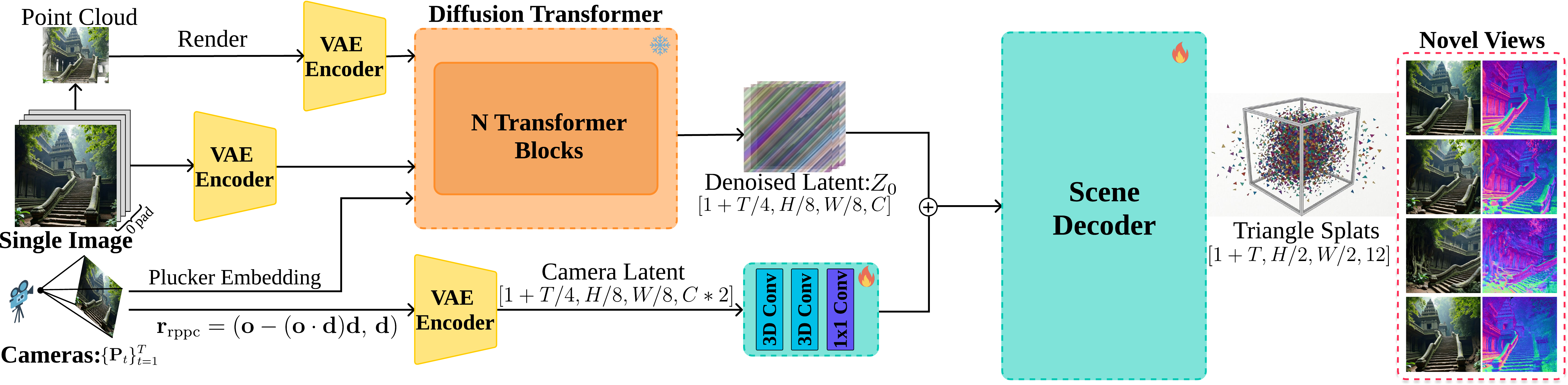}
    \caption{\textbf{Pipeline:} Starting from a single image, we construct a point-cloud-based control video by rendering along the target camera trajectory. The control video and camera embeddings condition a frozen video diffusion model \cite{cao2025uni3c}. The scene decoder then fuses denoised video latent with the camera latent and decodes triangle-splat scene representation for novel-view synthesis.}
    \label{fig:pipeline}
\end{figure*}

Given a single RGB image $\mathbf{I}_0 \in \mathbb{R}^{H \times W \times 3}$ and a camera trajectory $\{\mathbf{P}_t\}_{t=1}^{T}$, where each  $\mathbf{P}_t = (\mathbf{K}_t, \mathbf{R}_t, \mathbf{t}_t)$ denotes camera intrinsics and extrinsics, our goal is to produce an explicit 3D scene representation that can be rendered from arbitrary viewpoints in real time. The scene is represented as a set of surface primitives in world space, decoded in a single forward pass.

\subsection{Pipeline Overview}
Our method augments a frozen camera-conditioned latent video diffusion model \cite{cao2025uni3c} with a feedforward scene decoder, as illustrated in \Cref{fig:pipeline}. At test time, the pipeline takes as input a single RGB image $\mathbf{I}_0$ and a target camera trajectory $\{\mathbf{P}_t\}_{t=1}^{T}$. Conditioned on the input view and camera information, the video generator outputs a denoised latent $\mathbf{z} \in \mathbb{R}^{F' \times C' \times H' \times W'}$. We train a scene decoder that maps the latent directly to explicit scene parameters. The decoder predicts a set of surface primitives, which are then converted into world coordinates to form the final scene representation. In this way, \methodname reuses the strong generative prior of the video model, enabling plausible generation of scene content beyond the input view in a single forward pass without expensive per-scene optimization.

\subsection{Feedforward Triangle Splatting}
We represent the scene as a set of triangle splats, following differentiable triangle rendering~\cite{triangle_splatting}. Each triangle $\mathbf{T_m}$ is defined by three vertices $\mathbf{v}_i{\in} \mathbb{R}^3$, a color $\mathbf{c_m}$, a smoothness parameter $\sigma_m$ and an opacity $o_m \in [0,1]$.
To render a triangle, we project each vertex to the image plane with a standard pinhole camera model. Given camera intrinsics $\mathbf{K}$ and pose $(\mathbf{R}, \mathbf{t})$, the projected vertices are
\begin{equation}
    \mathbf{q}_{m,i} = \mathbf{K}(\mathbf{R_i}\mathbf{v}_{m,i} + \mathbf{t_i}) \, ,
\end{equation}
where the three points $\mathbf{q}_{m,i} \in \mathbb{R}^2$ form the projected triangle $T_m^{2\mathrm{D}}$ in the image plane. To enable differentiable rasterization, we assign each pixel $p$ a soft coverage value $I_m(p) \in [0,1]$ via a window function described below. The rendered image is then obtained by accumulating the contributions of all overlapping triangles in front-to-back depth order, following the standard alpha-compositing equation used in differentiable splatting methods \cite{2dgs, 3dgs, held20253d}.

\begin{figure*}[t]
    \centering
    \includegraphics[width=\textwidth]{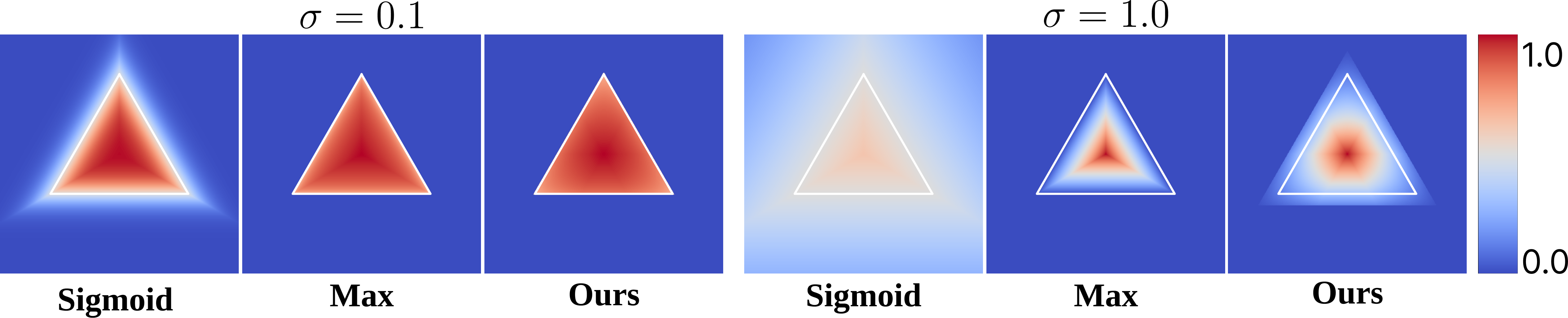}
    \caption{\textbf{Window Function:} Comparison of sigmoid-based window function \cite{held20253d, deng2020cvxnet}, max edge distance is used in \cite{triangle_splatting} and ours. \methodname function extends the influence outside the triangle boundary and improves gradient flow by routing to all three vertices.}
    \label{fig:window_function}
\end{figure*}

\paragraph{Decoding Triangles:}

A triangle splat can be parametrized by the model in several ways -- directly predicting 3D vertices, edge vectors, or a canonical template with learned scale and rotation. Because triangles are flat primitives, orientation errors can yield negligible contributions to the rendered image, whereas degenerate solutions, such as three vertices forming a line, degrade training stability and require additional constraints. Thus, feedforward training is particularly sensitive to the choice of parameterization. We predict each triangle relative to a ray-centered local frame and convert it to world space during post-processing. Concretely, each decoder output token predicts the parameters of a single triangle splat for a local $2\times2$ image region. For an anchor ray with origin $\mathbf{r}_o$ and direction $\mathbf{r}_d$, the network predicts a depth value $D$, three shape parameters, rotation parameters, color coefficients, opacity, and the sharpness parameter $\sigma$. The triangle center is placed at $\mathbf{r}_o + D \cdot \mathbf{r}_d$, while its in-plane geometry is first defined in a 2D coordinate system tangent to the ray.

Regressing three unconstrained vertices can result in degenerate triangles. Instead, we start from a canonical centered equilateral triangle in 2D and transform it with a lower-triangular matrix
\begin{equation}
    \mathbf{L} = \begin{bmatrix}
        L_{00} & 0 \\
        L_{10} & L_{11}
    \end{bmatrix} \in \mathbb{R}^{2 \times 2},
\end{equation}
whose coefficients are directly regressed by the model. The diagonal terms $L_{00}$ and $L_{11}$ are forced to be positive to maintain a valid triangle, while the off-diagonal term $L_{10}$ controls shear. Applying $\mathbf{L}$ to the canonical triangle yields a flexible family of anisotropic 2D triangles while guaranteeing strictly positive area and avoiding degenerate configurations during training. We then translate the transformed vertices so that their centroid coincides with the anchor point along the ray.

Finally, the local 2D triangle is lifted to 3D using a ray-tangent frame. Orientation is parametrized by two residual tilt angles and an in-plane spin angle. We found this decomposition to be more numerically stable than predicting a full 3D rotation, such as a quaternion, for each triangle. Early in training, direct prediction of world-space rotation often leads to unstable orientation, vanishing render support, and model divergence. By predicting residual rotations around a ray-aligned frame, triangles inherit position from the predicted ray depth, shape from the Cholesky-style 2D transform, and orientation from a locally constrained rotation. In our experiments, the rotation parameterization is a key ingredient for stable feedforward latent decoding of non-volumetric primitives.

\paragraph{Window Function:}
A window function replaces the hard triangle with a smooth approximation, enabling effective gradient flow. Thus, a choice of approximation is crucial for a stable feedforward decoder. For a projected triangle $T_m^{2\mathrm{D}}$, let $L_{m,i}(p) = \mathbf{n}_{m,i}^\top p + d_{m,i}$ denote the signed distance to the $i$-th supporting edge line, where the outward normals $\mathbf{n}_{m,i}$ are chosen such that $L_{m,i}(p) < 0$ inside the triangle. Let $s_m$ be the triangle incenter and let $\rho_m = -\max_i L_{m,i}(s_m)$ denote its inradius in screen space. We define the normalized edge response as
\begin{equation}
    u_{m,i}(p) = -\frac{L_{m,i}(p)}{\rho_m}
\end{equation}
so that $u_{m,i}(p) > 0$ inside the triangle and $u_{m,i}(p) = 1$ at the incenter. We then apply a shifted clipping
\begin{equation}
    r_{m,i}(p) = \mathrm{clamp}\left(u_{m,i}(p) + \epsilon,\, 0,\, 1\right),
\end{equation}
where $\epsilon > 0$ extends support beyond the exact boundary. The final window value is
\begin{equation}
    I_m(p) = \left(\prod_{i=1}^{3} r_{m,i}(p)\right)^{\sigma_m},
\end{equation}
where $\sigma_m$ controls the sharpness of the splat. Each pixel receives a signal from the full triangle rather than only the most active edge. The shift by $\epsilon$ also yields non-zero derivatives beyond the boundary, which improves stability early in training.

Compared with the original triangle-splatting formulation~\cite{triangle_splatting}
our formulation avoids the max reduction inside the window, as shown in \Cref{fig:window_function}. In practice, this improves gradient flow, which is particularly important in our feedforward latent model.

\subsection{Feedforward Scene Decoder}
In this section, we describe our feedforward scene decoder that regresses the 3D scene parameters.
\paragraph{Architecture:}
In contrast to other feedforward scene generation methods \cite{lyra, wonderland} that train small transformer decoders \cite{dosovitskiy2020image} or mamba-based architectures \cite{gu2023mamba} from scratch, we modify the decoder of a pretrained video VAE instead. Concretely, we reuse the RGB decoder backbone of Wan-2.1 \cite{wan}. We introduce camera conditioning via zero-convolutional blocks and attach lightweight output heads that map intermediate decoder features to triangle parameters rather than RGB pixels. In addition, we remove the last upsampling stage of the decoder to reduce the number of predicted primitives, predicting triangle parameters for a $2\times2$ pixel area. This transfer-learning setup simplifies optimization of the challenging problem: the pretrained decoder implicitly captures local appearance and spatial patterns,  allowing the model to focus on high-quality rendering and accurate geometry. 

\paragraph{Camera Conditioning:}
We encode camera information as dense per-pixel ray embeddings aligned with the video latent. Starting from the pixel-aligned Pl\"ucker ray embedding
\begin{equation}
    \mathbf{r}_{\mathrm{pl}} = (\mathbf{o} \times \mathbf{d},\, \mathbf{d}),
\end{equation}
where $\mathbf{o} \in \mathbb{R}^3$ and $\mathbf{d} \in S^2$ are the ray origin and direction, we follow DiffusionGS~\cite{diffusiongs} and replace the moment vector with the point on the ray closest to the world origin:
\begin{equation}
    \mathbf{r}_{\mathrm{rppc}} = (\mathbf{o} - (\mathbf{o} \cdot \mathbf{d})\mathbf{d},\, \mathbf{d}).
\end{equation}
This RPPC parameterization better exposes the ray position and relative depth to the decoder.

Let $\mathbf{r}^{\mathrm{rppc}} \in \mathbb{R}^{T \times 6 \times H \times W}$ denote the dense RPPC maps for a video. Following \cite{lyra}, we split them into reference-point and direction components, $\mathbf{r}^{\mathrm{ref}}, \mathbf{r}^{\mathrm{dir}} \in \mathbb{R}^{T \times 3 \times H \times W}$, and encode them separately with the pretrained VAE encoder $\mathcal{E}$:
\begin{equation}
    \mathbf{E}^{\mathrm{ref}} = \mathcal{E}(\mathbf{r}^{\mathrm{ref}}),
    \qquad
    \mathbf{E}^{\mathrm{dir}} = \mathcal{E}(\mathbf{r}^{\mathrm{dir}}),
\end{equation}
where $\mathbf{E}^{\mathrm{ref}}, \mathbf{E}^{\mathrm{dir}} \in \mathbb{R}^{T' \times C \times H' \times W'}$ match the video autoencoder downsampling. We concatenate them along channels and project back to the decoder width:
\begin{equation}
    \mathbf{E}^{\mathrm{cam}} = \phi\left(\left[\mathbf{E}^{\mathrm{ref}}; \mathbf{E}^{\mathrm{dir}}\right]\right),
    \qquad
    \mathbf{E}^{\mathrm{cam}} \in \mathbb{R}^{T' \times C \times H' \times W'},
\end{equation}
where $\phi$ is a lightweight learned fusion layer. We inject $\mathbf{E}^{\mathrm{cam}}$ through zero-initialized blocks, so that camera features remain aligned with visual latents as the model gradually learns to use them. Because the decoder is time-causal, we train on shorter sequences and still decode larger scenes during inference.

\subsection{Model Training}
We use a pre-trained video diffusion model and only train the scene decoder. Our training relies on a dataset of videos with known camera trajectories as well as depth and normal maps. For each video, we precompute its latents using the frozen VAE. The scene decoder is then trained to regress the 3D scene representation from the video latents and the camera trajectory.

\paragraph{Implementation Details.}
We use Uni3C \cite{cao2025uni3c} as the video model, which is built on top of Wan-2.1 \cite{wan} and generates 49 to 81 frames using a resolution of $432\times768$. The VAE encoder temporarily downsamples the video by a factor of $r_t=4$ and spatially by $r_s=8$. We train the scene decoder in four progressive stages from $320$ to $768p$, due to the high computational cost. Depending on the stage, a total of $V = N$ supervision views are used, equally split between seen and unseen views. More details are presented in \Cref{sec:details}. We use the AdamW optimizer with a learning rate of $1e-4$ and train our model on 8 H100 GPUs for 200 000 iterations.

\paragraph{Losses:}
\methodname is supervised with a combination of photometric and geometry losses, together with several regularization terms. In line with other feedforward 3D models \cite{longlrm, lyra, wonderland}, we use a pixel-wise $L_2$ loss, along with a perceptual LPIPS loss \cite{perceptual_loss}, between the rendered and target frames. We also supervise rendered depth with a scale-invariant disparity loss, as in MiDaS \cite{midas}. Finally, we directly supervise our rendered normals against the pseudo-ground-truth normals $\mathcal{L}_\mathrm{N} = \frac{\sum_{i} M_i \left( 1 - \hat{\mathbf{n}}_i \cdot \mathbf{N}_i \right)}{\sum_i M_i}$, where $M_i = 1$ if $\alpha_i > 0.5$, $\hat{\mathbf{n}}$ is rendered normal and $N$ is a ground truth.
% The normals regularization term thus encourages the predicted surface orientation to align with the pseudo-ground-truth normals at valid pixels. 
Finally, during the high-resolution training stage, we apply an opacity regularization term, as commonly used in feedforward 3D Gaussian methods \cite{longlrm, lyra}, and remove triangles with opacity below $40\%$. The full objective is a weighted sum of these terms:
\begin{equation}
    \mathcal{L} = \lambda_{\mathrm{rgb}} \mathcal{L}_{2}
    + \lambda_{\mathrm{perc}} \mathcal{L}_{\mathrm{LPIPS}}
    + \lambda_{\mathrm{D}} \mathcal{L}_{\mathrm{D}}
    + \lambda_{\mathrm{N}} \mathcal{L}_{\mathrm{N}}
    + \lambda_{\mathrm{O}} \mathcal{L}_{\mathrm{O}},
\end{equation}
where $\lambda_{\mathrm{rgb}}=1.0$, $\lambda_{\mathrm{perc}}=0.5$, $\lambda_{\mathrm{D}}=0.01$, $\lambda_{\mathrm{N}}=0.01 $ and $\lambda_{\mathrm{O}} = 0.001$.

\subsection{Opaque Mesh Conversion:}
For a game-engine-compatible format, we use the global triangle sharpness $\sigma$ and the connected-triangles support, following \cite{held2025meshsplatting}. We set the initial $\sigma=0.5$ and convert the predicted semi-opaque and sharp triangles into a mesh using a fast post-processing optimization over generated video frames. This procedure converts the semi-transparent renderable soup into a more coherent, opaque triangle, allowing direct export to various rendering engines. Starting from the feedforward output, we first refine depth, geometry, color, and opacity under the same photometric rendering objective used during training, then perform 50 iterations of an aggressive opacity-selection stage that pushes per-triangle opacity toward binary values and removes triangles with low support. The surviving triangles are snapped to near-opaque opacity, locally densified near boundaries, and stitched by merging mutually nearest boundary vertices and pruning floaters. Last, we run a brief repair stage that adjusts vertex positions and colors to recover image fidelity after the topology change.

%% file: sections/4-evaluation.tex
\section{Evaluation}

We evaluate \methodname on the task of feedforward 3D scene generation from a single input image. Since most methods we compare with are closed-source, we follow the evaluation protocol described in Lyra \cite{lyra}, Bolt3D \cite{szymanowicz2025bolt3d}, and Wonderland \cite{wonderland}.
%Moreover, to fairly analyze impact of representation and isolate it from pipeline differences, we train comparable 3D Gaussian Splatting (3DGS) and 2D Gaussian Splatting (2DGS) variants under the same training hyperparameters and evaluate them under the same protocol. 

\paragraph{Dataset:}
We train on a mixture of real and synthetic videos. Real videos from RealEstate10K \cite{re10k} and DL3DV \cite{dl3dv} provide realistic scene statistics, appearance variation, and naturally occurring camera motion. However, the camera trajectories from SfM are often noisy and scale-ambiguous. For these datasets, we use the camera annotations from RealCam-Vid \cite{realcam}. Synthetic data complements real videos in two ways. First, we sample $25{,}000$ images from the object-centric S3OD \cite{kupyn2026s3od} dataset, synthesize videos with the Uni3C model \cite{cao2025uni3c} with basic camera motions such as pans and zooms, and store the corresponding trajectory metadata. This significantly expands the data distribution and covers a wider variety of scenes in contrast to a limited set of real videos. Second, we regenerate videos from the first frame and the target trajectory for RealEstate10K and DL3DV. These regenerated sequences ensure that the model learns from the actual distribution of the video diffusion model and adapts to its noise and biases, reducing the train--test gap when decoding from its latent outputs. In practice, we first pretrain on the larger synthetic data and then perform a final finetuning stage on the real videos. To map all scenes to the same scale, we predict metric camera poses and depths with MapAnything \cite{keetha2025mapanything}. We observed that Uni3C does not perfectly match the input camera conditions for challenging trajectories, so we recomputed the camera trajectory using MapAnything predictions for the generated videos. For real videos, we keep RealCam-Vid poses rescaled to metric scale. Pseudo-ground-truth surface normals are predicted by NormalCrafter \cite{bin2025normalcrafter}.

\begin{figure*}[t]
    \centering
    \includegraphics[width=\textwidth]{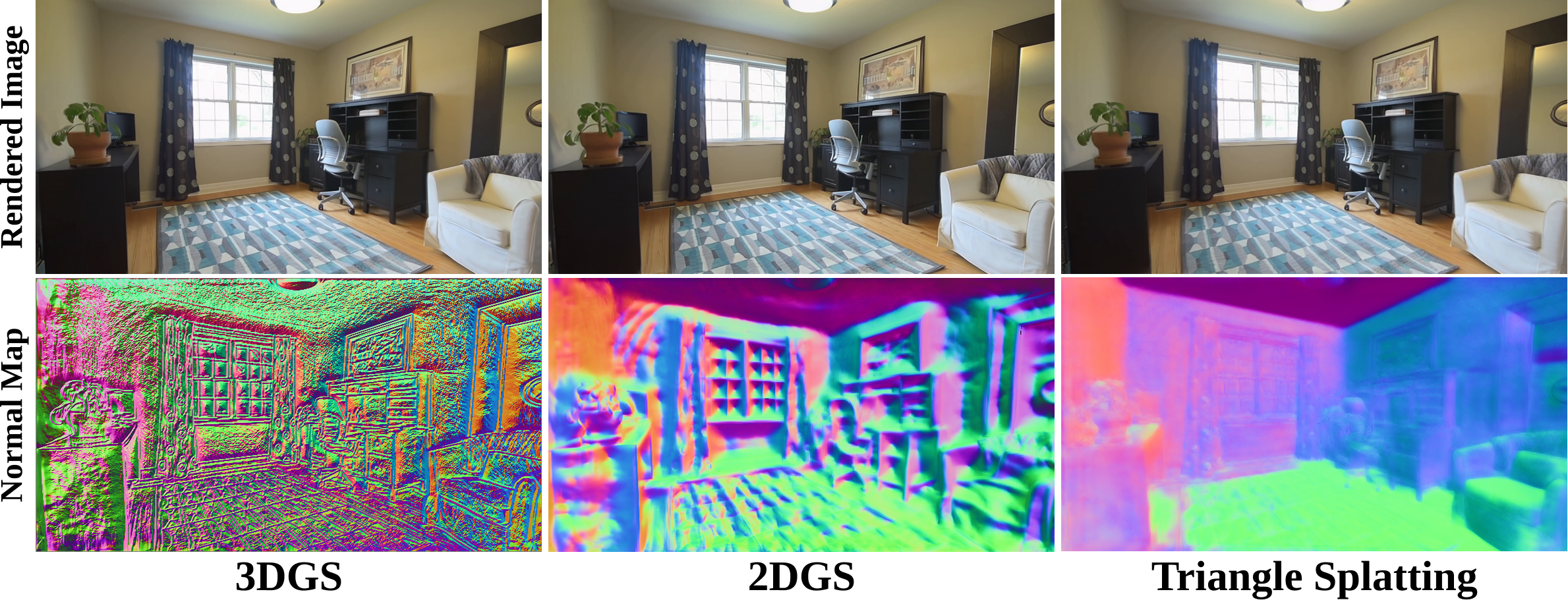}
    \caption{\textbf{Geometric Quality:} The latent triangle model generates finer, more accurate geometry compared to Gaussian representations that are optimized for visual quality, while still maintaining high rendering fidelity.}
    \label{fig:geo_quality}
\end{figure*}

\subsection{Scene Generation Evaluation}

\input{tables/scene_generation}
We evaluate our approach for the image-to-3D scene generation task on RealEstate10K and DL3DV datasets in Table~\ref{tab:scene_comparison}.
To fairly analyze the impact of representation and isolate it from pipeline differences, we train comparable 3D Gaussian Splatting (3DGS) and 2D Gaussian Splatting (2DGS) variants with the same training hyperparameters and evaluate them under the same protocol. We also include state-of-the-art methods based on 3DGS representation for comparison.

In addition to the standard image quality metrics, we also evaluate the geometric accuracy of the generated scene by directly comparing rendered normal maps with normals extracted from ground-truth frames. Since \methodname is directly supervised with NormalCrafter \cite{bin2025normalcrafter}, we employ Metric3D-v2 \cite{hu2024metric3dv2} to lower the impact of model bias. For the 3DGS variant, we compute normal with finite differences from nearby depth points \cite{2dgs}. Since 3DGS are volumetric blobs, they do not define clear geometry, generating near-random normals. Importantly, 2DGS explicitly models surfaces, yet it cannot be effectively supervised with high-quality normals. In our experiments, the direct supervision leads to numerical instability and model divergence; thus, we train with the original objective of normal self-consistency \cite{2dgs}. This improves geometric quality over 3DGS, yet the predicted surfaces remain soft and less structured than those produced by triangle splatting. The triangle model achieves superior geometric quality \Cref{fig:geo_quality}, with a cosine similarity of 0.853 to Metric3D labels, compared to 2DGS's 0.587 (averaged over both RealEstate10K and DL3DV). At the same time, visual metrics are comparable to those of other state-of-the-art methods.
 \Cref{tab:scene_comparison} also demonstrates the overall effectiveness of our training pipeline. All three representations generate high-quality visuals comparable to or superior to current state-of-the-art methods. 3DGS remains the strongest rendering-oriented baseline overall due to its volumetric nature, improving the quality of novel view synthesis over previous state-of-the-art methods, thus serving as an approximate upper bound for the triangle splats. Essentially, its blob-like parameterization is easier to predict, can naturally handle various 3D structures, such as semi-opaque fog and thin edges via "needle"-like Gaussians, and directly optimizes pixel-wise metrics such as PSNR. Triangles, instead, recover sharper, more geometrically faithful surfaces, provide an explicit, non-volumetric representation, and are substantially better aligned with downstream mesh extraction and real-time graphics pipelines. Overall, these results support explicit triangle-based feedforward scene decoding as a valid alternative when geometric accuracy and downstream compatibility matter.

\subsection{Mesh Conversion Evaluation}
Our approach \methodname provides the key benefit that predicted triangles can be converted into an opaque mesh with a lightweight post-processing step.
We compare the quality of this mesh with the meshes obtained from 2DGS and 3DGS representations in \Cref{tab:mesh_conversion}. Importantly, existing methods for surface extraction from 3DGS/2DGS rely on dense view coverage and are highly sensitive to hyperparameter choices. We observe that, given our sparse-view coverage and smaller scene scale, each scene requires careful hyperparameter tuning, and no single set of parameters works well across indoor and outdoor scenes. Thus, traditional marching cubes or TSDF surface-extraction methods simply fail in most scenes. In contrast, our predictions only require simple postprocessing, forcing opaque sharp triangles and connecting nearby edges, which significantly reduces the hyperparameter sensitivity. Consequently, the opaque meshes obtained via triangles achieve a PSNR improvement of over 7 dB compared to 3DGS meshes on RealEstate10K.

\input{tables/mesh_conversion}

\subsection{Ablation Study}

\input{tables/main_ablation}

We ablate the main design choices of \methodname across parameterization, rendering, conditioning, and post-processing. In particular, we study the effect of the ray-centered triangle parameterization, the modified triangle window function, the rotation parametrization, and model architecture. These ablations show that stable feedforward decoding of triangle primitives depends on the combination of all components. Predicting rotation directly in world space leads to model divergence to complete noise or empty renders. Employing the LongLRM Mamba-based decoder used in Lyra also underperforms, suggesting that its limited capacity is insufficient for decoding complex non-volumetric primitives. Changing the predicted parameterization reduces training stability, and reverting to the original window formulation weakens gradient flow.

%% file: tables/scene_generation.tex
\begin{table}[t]
\centering
\scriptsize
\caption{\textbf{Novel View Synthesis and Geometry Quality.} Feedforward triangle splatting generates significantly more accurate geometry compared to other representations while maintaining high visual quality compared to state-of-the-art methods. Our 3DGS variant achieves the highest visual fidelity, confirming the effectiveness of the training pipeline and design choices. Geometric quality denotes accuracy of generated normals. We report mean geometric quality over both RealEstate10K and DL3DV.}
\label{tab:scene_comparison}
\setlength{\tabcolsep}{3.5pt}
\renewcommand{\arraystretch}{1.08}
\resizebox{\linewidth}{!}{%
\begin{tabular}{l l ccc ccc cc}
\toprule
\multirow{2}{*}{\textnormal{Method}} & \multirow{2}{*}{\textnormal{Representation}}
& \multicolumn{3}{c}{\textnormal{RealEstate10K}}
& \multicolumn{3}{c}{\textnormal{DL3DV}}
& \multicolumn{2}{c}{\textnormal{Geometry Quality}} \\
\cmidrule(lr){3-5} \cmidrule(lr){6-8} \cmidrule(lr){9-10}
& & PSNR~$\uparrow$ & SSIM~$\uparrow$ & LPIPS~$\downarrow$
& PSNR~$\uparrow$ & SSIM~$\uparrow$ & LPIPS~$\downarrow$
& $L_1\,\downarrow$ & Cosine$\,\uparrow$ \\
\midrule
ZeroNVS     & 3DGS      & 13.01 & 0.378 & 0.448 & 13.35 & 0.339 & 0.465 & -- & -- \\
ViewCrafter & 3DGS      & 16.84 & 0.514 & 0.341 & 15.53 & 0.525 & 0.352 & -- & -- \\
Wonderland  & 3DGS      & 17.15 & 0.550 & 0.292 & 16.64 & 0.574 & 0.325 & -- & -- \\
Bolt3D      & 3DGS      & 21.54 & 0.747 & 0.234 & -     & -     & -     & -- & -- \\
Lyra        & 3DGS      & 21.79 & 0.752 & 0.219 & 20.09 & 0.583 & 0.313 & -- & -- \\
\midrule
\textbf{\methodname} & 3DGS      &   \textbf{22.39}    &   \textbf{0.762}    &  \textbf{0.203}     &   \textbf{20.71}    &   \textbf{0.663}    &   \textbf{0.275}    & 0.686 & 0.116 \\
\textbf{\methodname} & 2DGS      &   22.03    &   0.734    &   0.219    &   20.44    &   0.647    &   0.284    & 0.388 & 0.587 \\
\textbf{\methodname} & Triangles &   21.45    &  0.710    &   0.245    &   20.04    &   0.627    &   0.314    & \textbf{0.211} & \textbf{0.853} \\
\bottomrule
\end{tabular}
}
\end{table}

%% file: tables/mesh_conversion.tex
\begin{table}[htb]
\centering
\small
\caption{\textbf{Opaque mesh conversion evaluation.} We compare opaque mesh extraction strategies across scene representations on RealEstate10K and DL3DV.}
\label{tab:mesh_conversion}
\setlength{\tabcolsep}{5pt}
\renewcommand{\arraystretch}{1.08}
\begin{tabular}{l l c ccc ccc}
\toprule
\multirow{2}{*}{\textnormal{Representation}} & \multirow{2}{*}{\textnormal{Conversion}} & \multirow{2}{*}{\textnormal{Vertices}}
& \multicolumn{3}{c}{\textnormal{RealEstate10K}}
& \multicolumn{3}{c}{\textnormal{DL3DV}} \\
\cmidrule(lr){4-6} \cmidrule(lr){7-9}
& & & PSNR~$\uparrow$ & SSIM~$\uparrow$ & LPIPS~$\downarrow$
& PSNR~$\uparrow$ & SSIM~$\uparrow$ & LPIPS~$\downarrow$ \\
\midrule
2DGS      & TSDF    & 5M & 15.89     &    0.633   &  0.468     &   12.00    &   0.433    &    0.563   \\
3DGS      & GS2Mesh \cite{wolf2024gs2mesh} & 4M & 14.18    &  0.619    &   0.452    &   12.31    &   0.465    &   0.541    \\
Triangles & Ours    & 0.5M &  \textbf{21.23}    &   \textbf{0.749}    &    \textbf{0.388}   &   \textbf{19.71}    &   \textbf{0.609}    &   \textbf{0.466}  \\
\bottomrule
\end{tabular}
\end{table}

%% file: tables/main_ablation.tex
\begin{table}[htb]
\centering
\small
\caption{\textbf{Ablation studies.} We analyze the effects of architecture, window function, and representation design on RealEstate10K and DL3DV.}
\label{tab:ablation}
\setlength{\tabcolsep}{3pt}
\renewcommand{\arraystretch}{1.05}
\resizebox{\linewidth}{!}{%
\begin{tabular}{l l l l ccc ccc}
\toprule
\multirow{2}{*}{\textnormal{Architecture}} & \multirow{2}{*}{\textnormal{Window Function}} & \multirow{2}{*}{\textnormal{Representation}} & \multirow{2}{*}{\textnormal{Rotation}}
& \multicolumn{3}{c}{\textnormal{RealEstate10K}}
& \multicolumn{3}{c}{\textnormal{DL3DV}} \\
\cmidrule(lr){5-7} \cmidrule(lr){8-10}
& & & & PSNR~$\uparrow$ & SSIM~$\uparrow$ & LPIPS~$\downarrow$
& PSNR~$\uparrow$ & SSIM~$\uparrow$ & LPIPS~$\downarrow$ \\
\midrule
Ours     & Ours & Ours & Global   & $< 10$           & $< 0.4$          & $> 0.4$          & $< 10$           & $< 0.4$          & $> 0.4$ \\
Ours     & Ours & 3 Offsets & Residual &   20.09          &   0.674          &   0.289          & 19.18           &   0.588          &   0.372 \\
Ours     & Triangle Splatting & Ours & Residual &   20.65          &   0.693          &   0.282          & 19.75           &   0.610          &   0.341 \\
LongLRM \cite{longlrm}  & Ours & Ours & Residual &   21.24          &   0.701          &   0.275          & 19.74           &   0.608          &   0.355 \\
Ours     & Ours & Ours & Residual &   \textbf{21.45} & \textbf{0.710}   & \textbf{0.245}   & \textbf{20.04} & \textbf{0.627}   & \textbf{0.314} \\
\bottomrule
\end{tabular}
}
\end{table}

%% file: sections/5-conclusion.tex
\section{Conclusion}
We presented \methodname, a feedforward approach for generating 3D scenes from a single input image. Our method combines the strong generative prior of a frozen camera-conditioned latent video model with a lightweight scene decoder that predicts triangle splats directly in a single forward pass. This design avoids expensive per-scene optimization, while still enabling plausible generation beyond the input view and real-time rendering of the resulting scene representation.
At the representation level, we showed that non-volumetric surface primitives can be decoded from video latents when the parameterization and rendering formulation are chosen carefully. In particular, our ray-centered triangle parameterization and modified differentiable triangle-splatting formulation improve optimization stability and enable practical feedforward prediction of explicit surface structure at high resolution. We further showed that a short post-processing stage can convert the predicted triangle soup into a substantially more coherent opaque mesh while preserving visual quality.
Our results suggest that latent video generation models can serve not only as image synthesizers, but also as powerful priors for direct 3D scene prediction. We expect this to encourage further work on explicit geometrically accurate feedforward scene generation and tighter integration between controllable generative models and real-time 3D rendering.

%% file: sections/7-appendix.tex
\section{Pipeline Flexibility}

A useful property of \methodname{} is that it generates scene parameters from the denoised latent of base Wan-2.1 without modifying the latent space or finetuning the diffusion transformer. At inference time, one can simply add \methodname decoder or replace the standard VAE RGB decoder, while leaving the upstream video generator unchanged. As a result, any Wan-2.1 variant finetuned from the base model can produce explicit triangle-based scene geometry instead of RGB frames. This includes image-to-video, text-to-video, video-to-video, control or editing pipelines \cite{wanmove, wang2025ati}, as well as more real-time \cite{yuan2026helios, gu2026anyflow}, interactive \cite{yang2025longlive, mao2026yume1}, long / autoregressive \cite{an2026onestory, huang2026self, cui2025self} or world-consistent variants~\cite{du2026videogpa, an2026vggrpo, kupyn2025epipolar, worldstereo}.

\begin{figure*}[h]
    \centering
    \includegraphics[width=\textwidth]{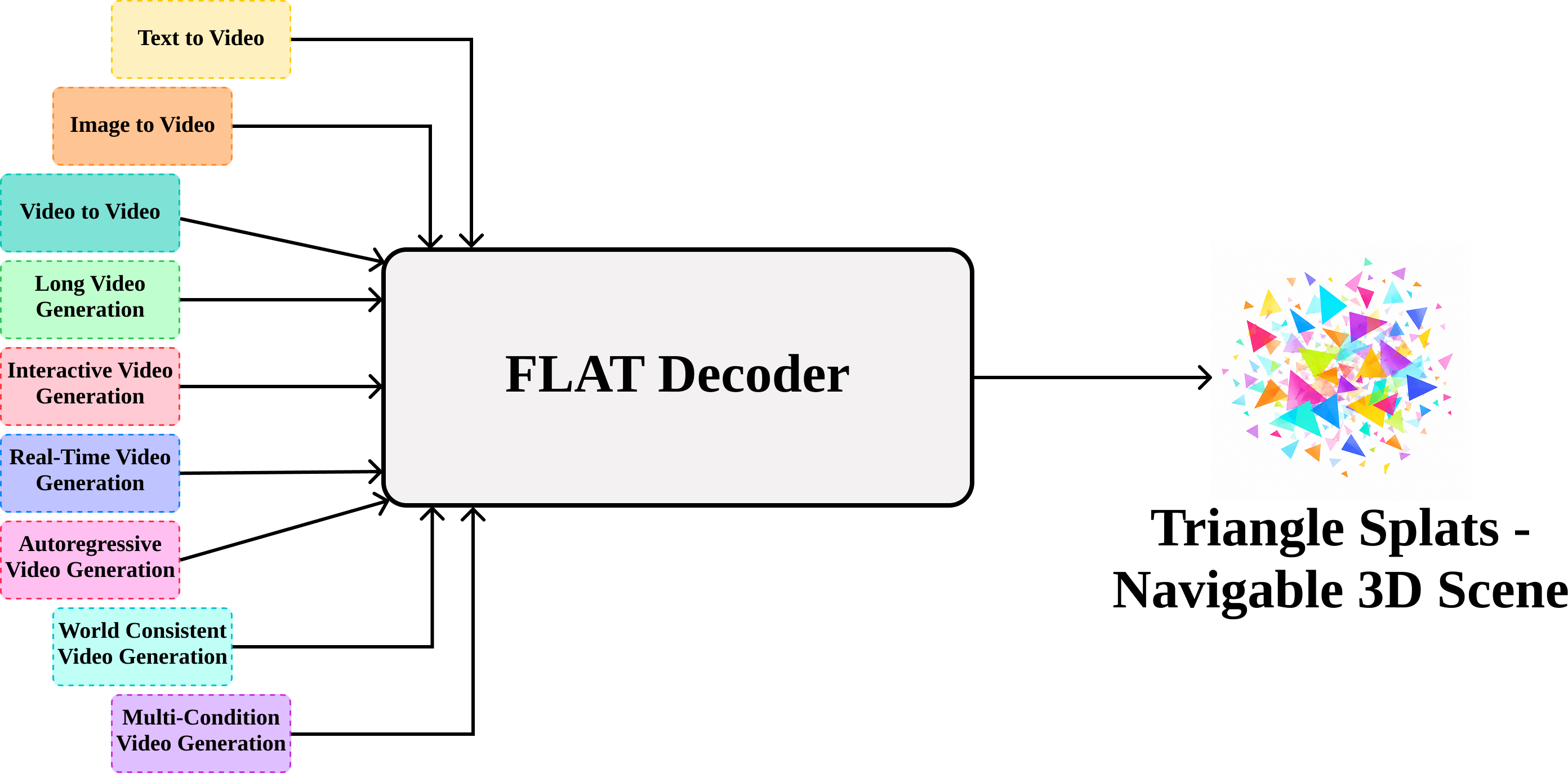}
    \caption{\textbf{Pipeline Flexibility:} \methodname{} replaces the standard RGB decoder with a latent scene decoder. Because multiple Wan variants share the same latent space, our scene decoder can be attached to any of these, including image-to-video, text-to-video, video-to-video, interactive, and world-consistent pipelines.}
    \label{fig:flexibility}
\end{figure*}

\begin{figure*}[h]
    \centering
    \includegraphics[width=\textwidth]{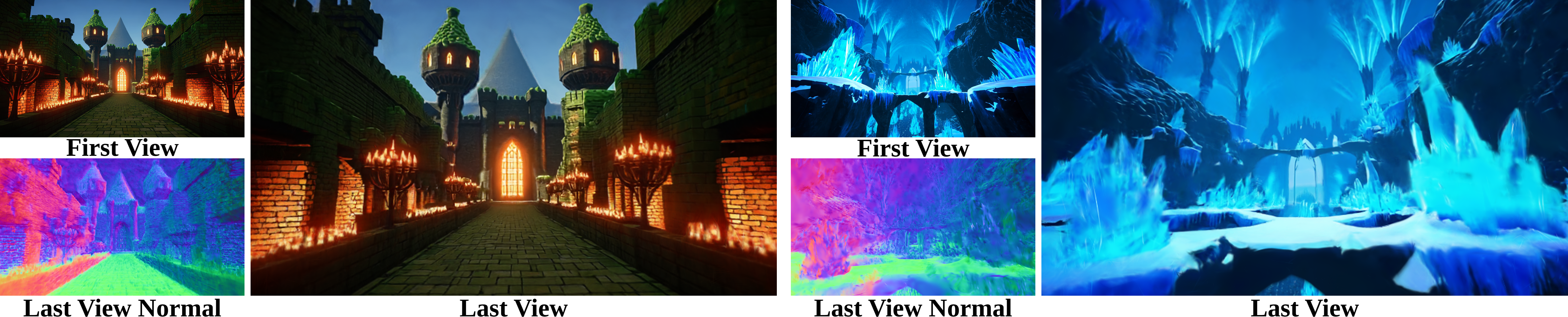}
    \caption{\textbf{Text-to-3D Scene:} Examples obtained by attaching \methodname{} to a Wan-2.1 text-to-video pipeline. For each scene, we show rendered views together with the corresponding predicted normal map. The examples demonstrate that the same latent scene decoder can convert text-to-video model latents into explicit geometry.}
    \label{fig:t2v}
\end{figure*}

This decoder-swap design makes \methodname{} practical for many applications. The method does not require a separate 3D decoder for each pipeline mode; instead, it reuses the shared latent representation. Consequently, improvements in the upstream generator, such as better motion quality, stronger conditioning, or new control interfaces, can be directly transferred to scene generation without retraining a separate scene decoder for each variant. In this sense, \methodname{} is best viewed as a geometry-aware explicit 3D generator for a broader family of video-generation pipelines. \Cref{fig:flexibility} illustrates this flexibility. We additionally verify this for text-to-video setting. \Cref{fig:t2v} shows two scenes produced from text prompts alone. In both examples, \methodname{} decodes the final text-to-video latents into explicit triangle-based scenes whose rendered views remain consistent with the predicted normals. These results suggest that the scene decoder is not limited to image-conditioned generation.

\section{Post Optimization}

Though \methodname{} is fully feedforward, a short test-time optimization can further improve both visual and geometric quality. The feedforward prediction already provides a strong initialization, so optimization mainly corrects common failure cases of the latent feedforward model, including surface misalignment, semi-transparent structures, floating low-importance triangles, thin objects and overly diffuse normal predictions. In practice, we find that even a very short refinement of as few as 250 steps is often sufficient to improve both visual and geometric quality.

\begin{figure*}[h]
    \centering
    \includegraphics[width=\textwidth]{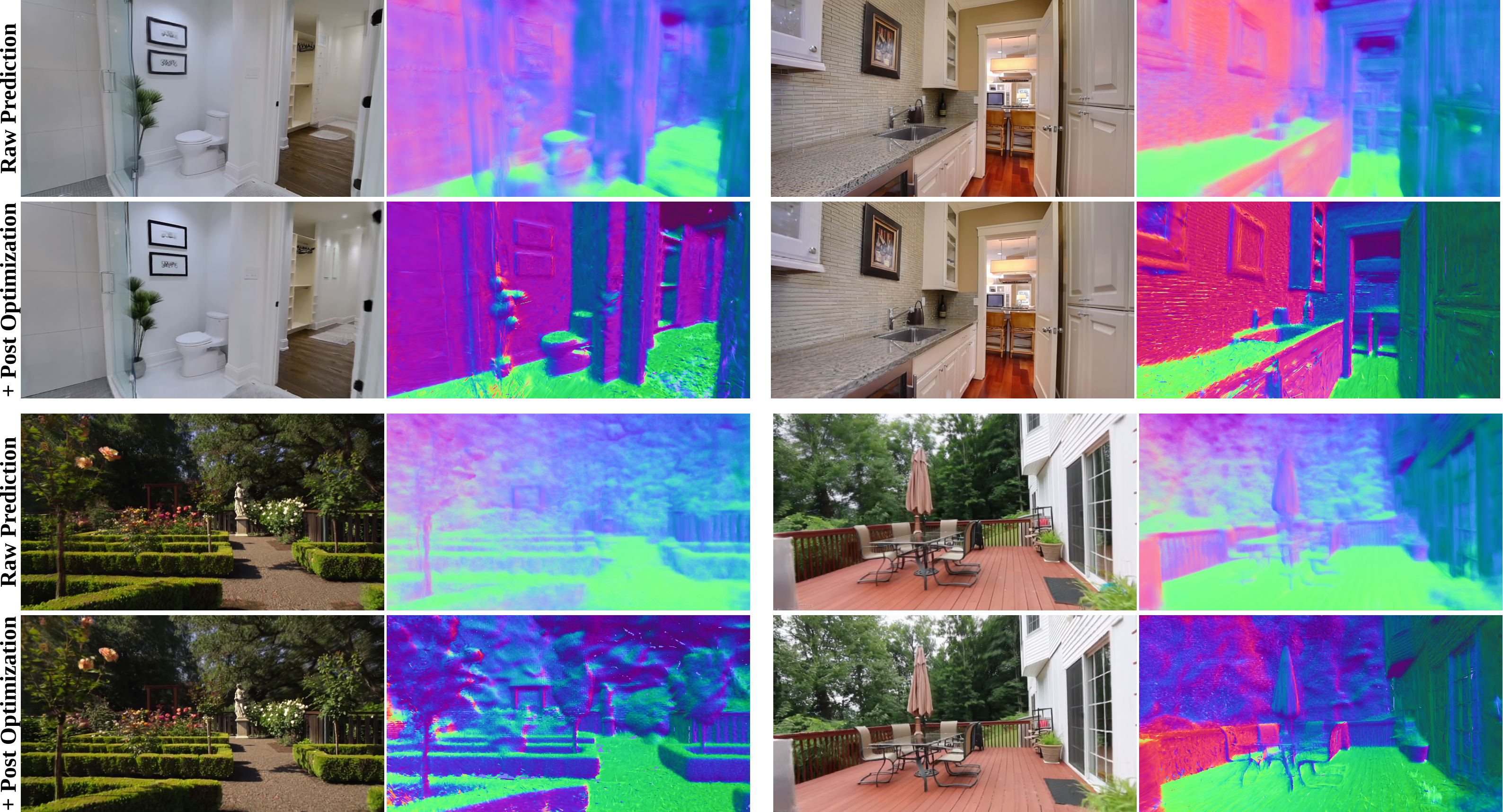}
    \caption{\textbf{Post Optimization:} Predictions and camera-space normals before and after optimization. A short refinement pass fixes common failures of the feedforward model and, together with aggressive pruning, produces cleaner geometry and sharper normals.}
    \label{fig:optimization}
\end{figure*}

The optional post-optimization further aligns trinalge splat renders with RGB frames. We apply aggressive pruning to remove weak or unsupported triangles. This final cleanup is especially important for geometric quality: while diffuse low-opacity triangles can partially hide local prediction errors in RGB space, they tend to blur surface orientation and soften normal boundaries. Removing them produces sharper normal maps and cleaner local geometry. Qualitative examples are shown in \Cref{fig:optimization}, where optimization improves both rendered appearance and predicted normals.

\begin{table}[h]
\centering
\caption{\textbf{Effect of Post Optimization on RealEstate10K.} A simple optimization pass consistently improves the feedforward prediction.}
\label{tab:post_optimization_re10k}
\begin{tabular}{@{}lccc@{}}
\toprule
\textbf{Method} & \textbf{PSNR} $\uparrow$ & \textbf{SSIM} $\uparrow$ & \textbf{LPIPS} $\downarrow$ \\ \midrule
\textbf{\methodname} & 21.45 & 0.710 & 0.245 \\
\textbf{\methodname{} + Optimization} & \textbf{23.01} & \textbf{0.790} & \textbf{0.230} \\ \bottomrule
\end{tabular}
\end{table}

\input{sections/6-limitations}

\begin{figure*}[h]
    \centering
    \includegraphics[width=\textwidth]{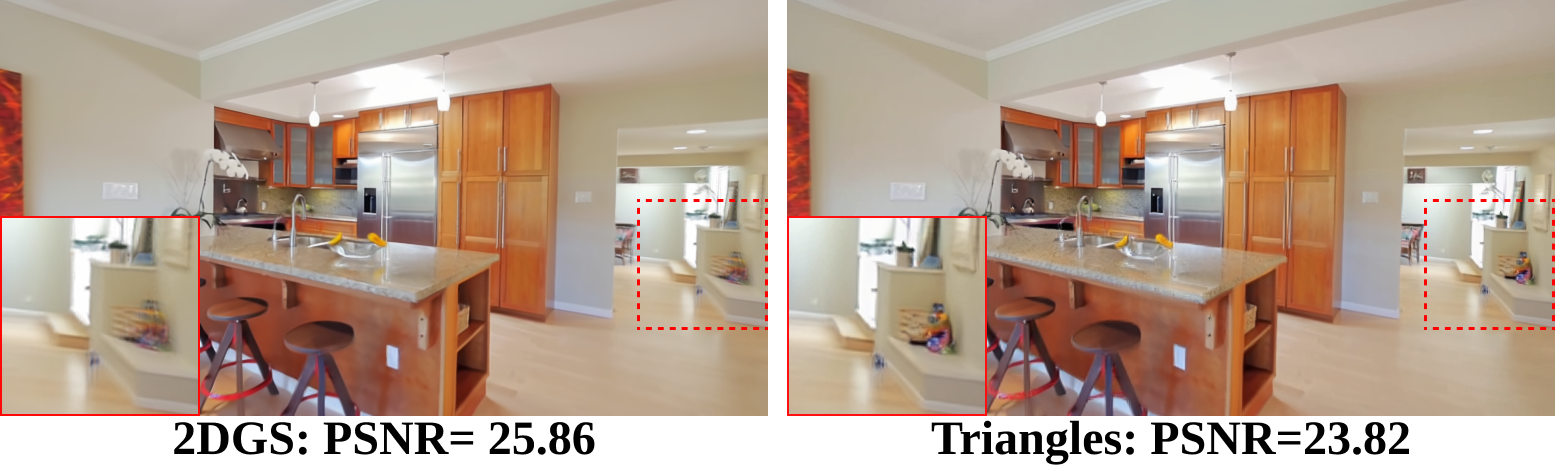}
    \caption{\textbf{Metric Limitations:} Gaussians are optimized for PSNR directly due to their inherent smoothness. The triangle model often generates sharper details while achieving lower PSNR.}
    \label{fig:psnr}
\end{figure*}

\begin{figure*}[h]
    \centering
    \includegraphics[width=\textwidth]{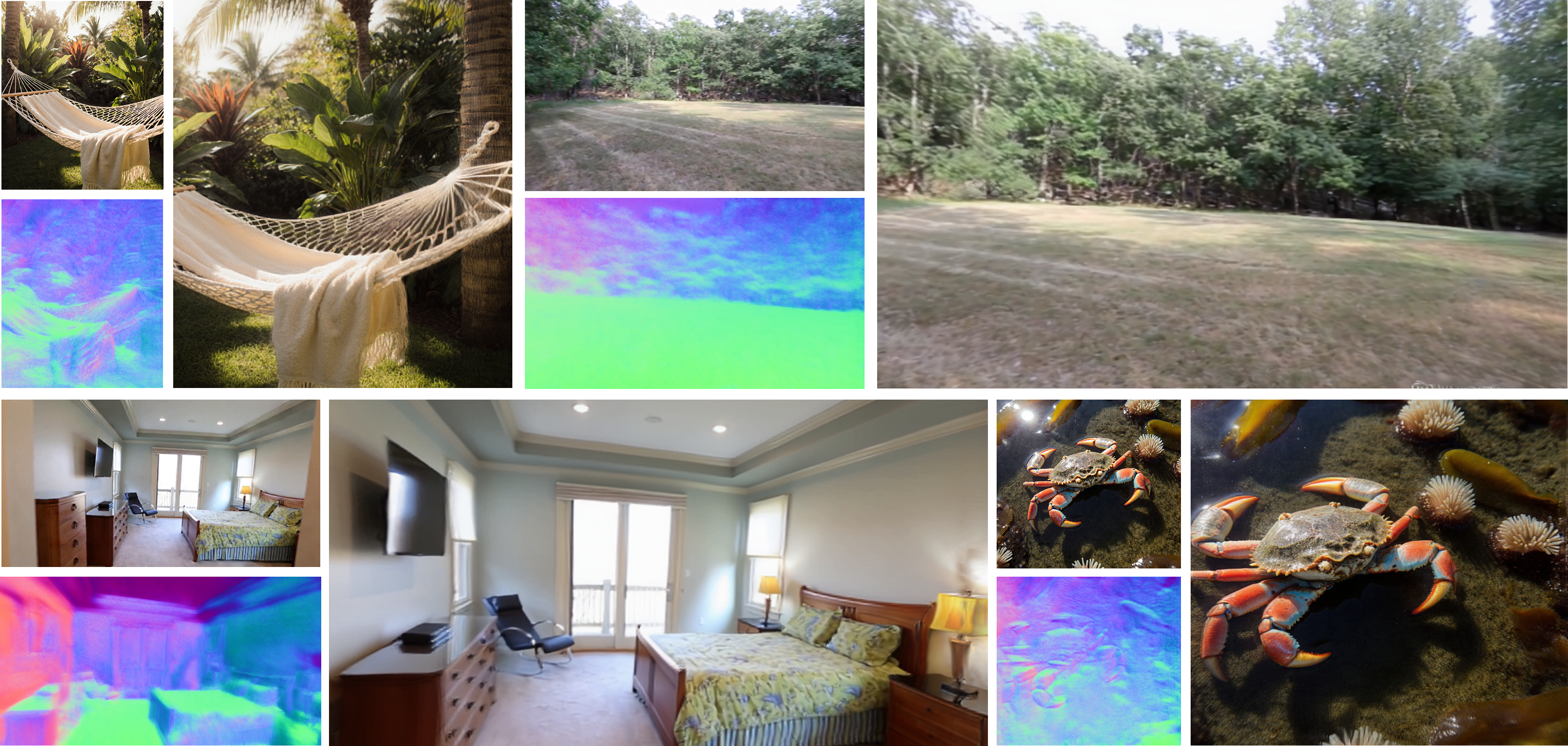}
    \caption{\textbf{Qualitative Results:} More qualitative results covering indoor, outdoor, and object-centric scenes, focusing on surface and visual quality. Each sample consists of input image, novel view and novel view normal map.}
    \label{fig:qual}
\end{figure*}

\begin{figure*}[h]
    \centering
    \includegraphics[width=\textwidth]{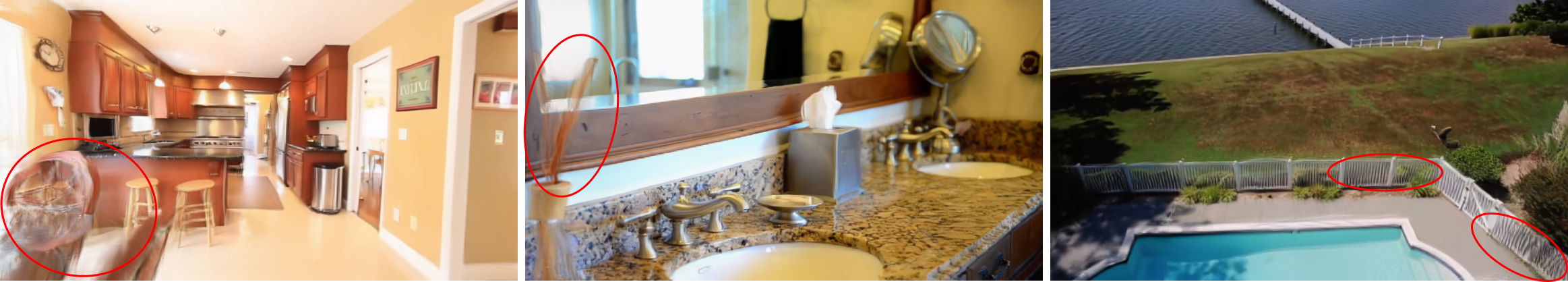}
    \caption{\textbf{Failure Cases:} Thin, elongated surfaces, tiny details and reflections remain challenging to model with triangles.}
    \label{fig:fail_cases}
\end{figure*}

\section{Training Details}
\label{sec:details}

\input{tables/training_details}

The multi-stage training pipeline follows a progressive resolution and view-scaling schedule across four stages \Cref{tab:training_schedule}. Stage 1 runs for $20,000$ iterations at $320\text{p}$ resolution, training on $17$-view RealEstate10K sequences, using $17$ input-conditioning views and $17$ target views to quickly adapt the decoder to a new task. Stage 2 trains on $49$-view trajectories from a mix of real and synthetic videos over $40,000$ iterations at $320\text{p}$ with $49$ target views sampled for supervision. Stage 3 increases the image resolution to $640\text{p}$ for $75,000$ iterations. Finally, Stage 4 performs high-resolution fine-tuning at $768\text{p}$ for $75,000$ iterations using memory-efficient 8-bit Adam optimization, pruning, and gradient checkpointing to reduce GPU memory consumption.

\begin{figure*}[h]
    \centering
    \includegraphics[width=\textwidth]{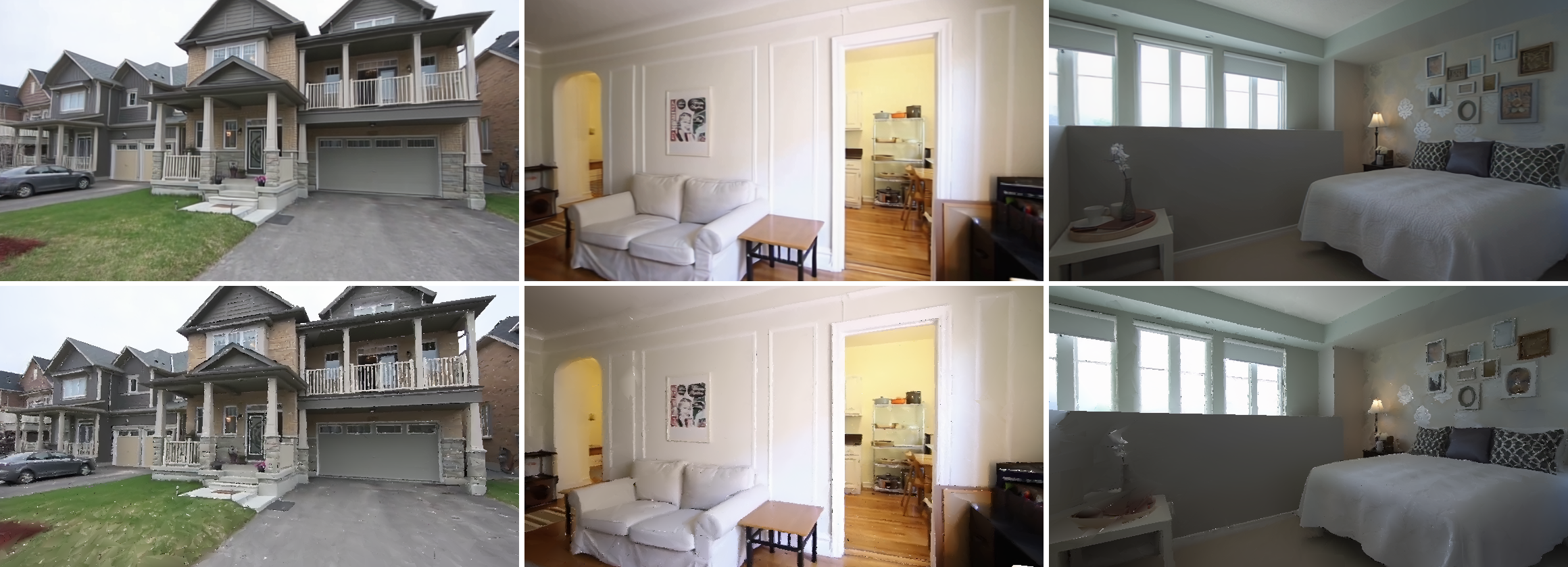}
    \caption{\textbf{Converted Mesh:} Top Row: semi opaque triangles predicted by the model. Bottow Row: opaque game engine compatible mesh generated by lightweight conversion step. The scene render remains high as strong semi-opaque geometrically accurate initial predictions simplify conversion process.}
    \label{fig:mesh_conversion}
\end{figure*}

\section{Scene Decoder Architecture}

\begin{table}[h]
\centering
\caption{{Computational Complexity:} Scene decoding consumes only a marginal fraction of the total compute relative to the video generation.}
\label{tab:performance_metrics}
\begin{tabular}{@{}lcccc@{}}
\toprule
\textbf{Resolution} & \textbf{Views} & \textbf{Parameters (M)} & \textbf{Compute (FLOPs)} & \textbf{Throughput (FPS)} \\ \midrule
768p & 49 & 73.34 & 9.8 TFLOPs & 2.70 \\ \bottomrule
\end{tabular}
\end{table}

Scene decoder matches Wan-2.1 VAE architecture. It utilizes 3D causal convolution ($\text{CausalConv3D}$) that pads the temporal dimension exclusively on the past frames to maintain temporal causality. The input video latents $x_v \in \mathbb{R}^{B \times 16 \times T' \times H' \times W'}$ and Pl\"{u}cker embeddings $x_p \in \mathbb{R}^{B \times 32 \times T' \times H' \times W'}$ are fused via a lightweight adapter, where $T'$, $H'$, and $W'$ denote the latent dimensions. To ensure the geometry conditioning does not destabilize the pre-trained latents at initialization, the final projection of the adapter is zero-initialized:

The fused latent $x_{in}$ is then processed by a 3D UNet-style decoder, outlined in Table~\ref{tab:wanscenedecoder}. Decoder employs RMSNorm, SiLU activations, and Scaled Dot-Product Attention (SDPA) throughout. Temporal upsampling occurs strictly in the first two upsampling stages, yielding a fixed $4\times$ temporal expansion. The final upsampling stage defaults to an identity pass, yielding an output tensor that is $2\times$ strided relative to the original image dimensions. \Cref{tab:performance_metrics} demonstrates detailed performance decoder metrics. We decode high resolution scene with 49 views in less than 300ms on H100 GPU.

\begin{table}[h]
\centering
\caption{Detailed architecture of the WanSceneDecoder. Output shapes are denoted as $(C, T', H', W')$ with the batch size $B$ omitted for brevity. Residual blocks (ResBlock) consist of RMSNorm, SiLU, and CausalConv3D layers.}
\label{tab:wanscenedecoder}
\resizebox{\linewidth}{!}{
\begin{tabular}{@{}lllc@{}}
\toprule
\textbf{Stage} & \textbf{Layer / Operation} & \textbf{Output Shape} & \textbf{Details} \\ \midrule
\textbf{Conditioning} 
& Pl\"{u}cker Adapter & $(16, T', H', W')$ & $3 \times 3 \times 3$ CausalConv, SiLU, ZeroConv \\
& Fusion ($x_{in}$) & $(16, T', H', W')$ & Addition with video latents \\ \midrule
\textbf{Input} 
& $\text{Conv}_{in}$ & $(384, T', H', W')$ & $3 \times 3 \times 3$ CausalConv, padding=1 \\ \midrule
\textbf{Mid Block} 
& $\text{ResBlock}_1$ & $(384, T', H', W')$ & Dropout=$0.0$ \\
& Attention & $(384, T', H', W')$ & SDPA, $1 \times 1$ Conv Projections \\
& $\text{ResBlock}_2$ & $(384, T', H', W')$ & Dropout=$0.0$ \\ \midrule
\textbf{Up Block 1} 
& $3 \times \text{ResBlock}$ & $(384, T', H', W')$ & \\
& Resample (3D) & $(384, 2T', 2H', 2W')$ & Nearest-exact, Temporal + Spatial up \\ \midrule
\textbf{Up Block 2} 
& $3 \times \text{ResBlock}$ & $(192, 2T', 2H', 2W')$ & \\
& Resample (3D) & $(192, 4T', 4H', 4W')$ & Nearest-exact, Temporal + Spatial up \\ \midrule
\textbf{Up Block 3} 
& $3 \times \text{ResBlock}$ & $(96, 4T', 4H', 4W')$ & \\
& Resample (Identity) & $(96, 4T', 4H', 4W')$ & Identity spatial pass, no upsampling \\ \midrule
\textbf{Output Head} 
& RMSNorm + SiLU & $(96, 4T', 4H', 4W')$ & Independent or Monolithic head \\
& $\text{Conv}_{out}$ & $(C_{out}, 4T', 4H', 4W')$ & $3 \times 3 \times 3$ CausalConv \\ 
& Reshape & $(4T' \cdot 4H' \cdot 4W', C_{out})$ & Flatten to Splat parameters \\ \bottomrule
\end{tabular}
}
\end{table}

\section{Mesh Conversion Analysis}

\begin{figure*}[h]
    \centering
    \includegraphics[width=\textwidth]{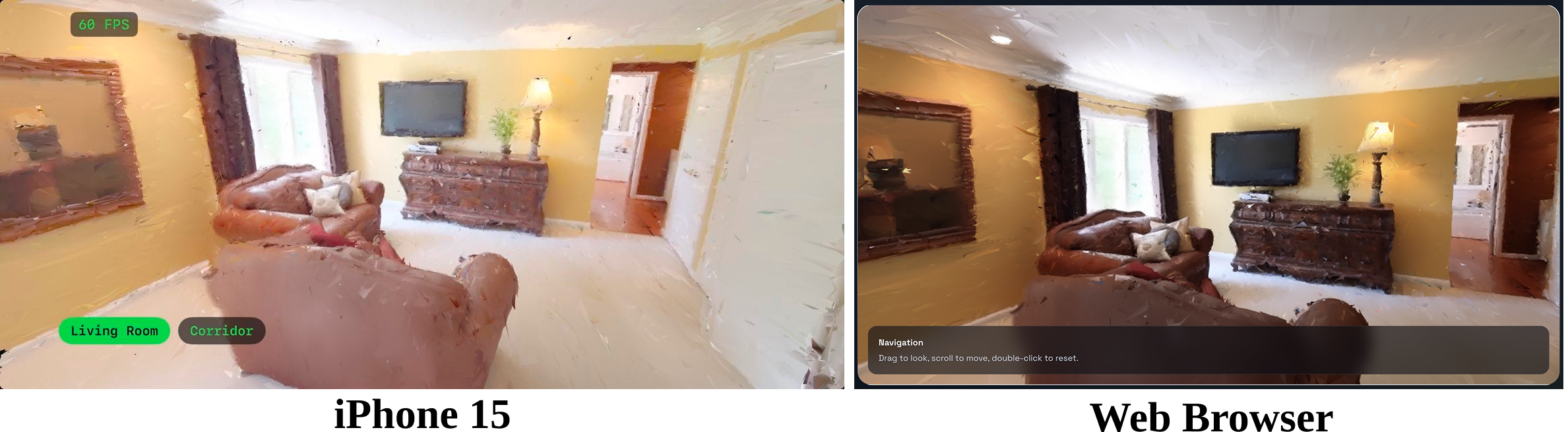}
    \caption{\textbf{Cross-Platform Rendering:} Rendering raw output without any postprocessing or mesh cleanup. The converted solid triangles can be rasterized by any rendering engine across various platforms, supporting high-resolution and high-fps efficient rendering across devices.}
    \label{fig:rendering}
\end{figure*}

To evaluate the quality of the conversion methods, we analyzed the geometry and topology of the output meshes. The direct conversion from soft triangles yields highly well-formed local geometry, with nearly zero degenerate faces $(0.02\%)$ and no fully isolated, disconnected triangles $(0.00\%)$. On average each triangle is connected to $3.1$ other triangles. This aligns with the expectation of 3 for regular manifold surfaces, proving ability to extract compact global structure. To quantify the remaining topological complexity, we measure the rate of non-manifold edges (edges with more than two connected faces), which accounts for $10.60\%$ of the mesh. Since the extracted mesh is not fully watertight, and rather represent a collection of locally connected surfaces, these non-manifold regions naturally emerge in locally dense zones where the network utilizes intersecting surface sheets to represent semi-transparent boundaries and fine details. The visual comparison is presented in \Cref{fig:mesh_conversion}. The resulting mesh can be effectively rendered on any platform with high efficiency. We have verified the compatibility by rendering our results in browser, on iPhone 15 and Google Pixel devices without incorporating any custom rendering engines \Cref{fig:rendering}.

%% file: sections/6-limitations.tex
\section{Limitations and Broader Impact}
\label{sec:limitations}

Despite strong geometric and visual quality \methodname still faces several limitations from the triangle representation and the feedforward generation. First, triangle splats are explicit, non-volumetric primitives, better aligned with surfaces but not optimized for standard novel-view synthesis performance. In particular, optimizing for pixel-level metrics such as PSNR remains more difficult than for 3DGS \Cref{fig:psnr}. Also, content such as very thin, elongated structures, reflections, semi-transparent regions, and fine, high-frequency details is challenging to model with triangles and remains a primary source of failure \Cref{fig:fail_cases}. Second, although our opaque-mesh conversion substantially improves usability, the resulting geometry is still sparser than a clean, watertight mesh: local connectivity can be incomplete, surfaces can remain oversharpened or fragmented, and producing dense, fully coherent geometry still requires additional post-processing. This is a limitation of all scene mesh recovery methods, so a clean, densely connected, watertight mesh remains an open problem.
The current model is also limited in scale. We train on a relatively small amount of data compared to modern video generation systems due to computational constraints and lack of high-quality ground truth data, and we expect both visual fidelity and geometric consistency to improve with dataset scaling. More fundamentally, \methodname predicts a scene from a single input image and one generated trajectory, so it must resolve severe 3D ambiguity from sparse view coverage. This can lead to incorrect geometry of occluded regions and failures on out-of-distribution scenes. In addition, our method currently targets a single generated scene or short camera path rather than a truly large explorable world \cite{worldstereo, shen2026lyra2}. Extending it to persistent large-scale environments requires integration with long world-consistent video generation.

In terms of broader impact, our approach can make 3D scene generation more practical for applications such as simulation, robotics, gaming, and AR/VR, where geometric structure matters alongside image quality. At the same time, the same technology could be misused to create realistic synthetic environments or deceptive media. As with other generative models, improving realism lowers the cost of producing misleading content. Finally, training and deploying such systems require substantial computing resources, which carry environmental costs.

%% file: tables/training_details.tex
  \begin{table}[h]
  \centering
  \caption{\textbf{Training Schedule:} The trainer progressively scales both number of views and resolution to reduce task complexity and improve computational efficiency}
  \label{tab:training_schedule}
  \begin{tabular}{lcccc}
  \toprule
  \textbf{Stage} & \textbf{Iterations} & \textbf{Resolution} & \textbf{Input Views} & \textbf{Target Views} \\
  \midrule
  1 & 20,000 & 320p & 17 & 17 \\
  2 & 40,000 & 320p & 49 & 49 \\
  3 & 75,000 & 640p & 49 & 49 \\
  4 & 75,000 & 768p & 49 & 24 \\
  \bottomrule
  \end{tabular}
  \end{table}